\documentclass[runningheads]{llncs}
\usepackage{graphicx}
\usepackage{amsmath,amssymb} 
\usepackage{overpic}
\usepackage[caption=false]{subfig}
\usepackage[breaklinks]{hyperref}
\usepackage{color}
\usepackage[vlined]{algorithm2e}
\begin{document}
\pagestyle{headings}
\mainmatter

\title{A Simple Approach to Intrinsic Correspondence Learning on Unstructured 3D~Meshes} 

\titlerunning{Correspondence Learning on Unstructured 3D~Meshes}

\authorrunning{I.~Lim, A.~Dielen, M.~Campen, L.~Kobbelt}

\author{Isaak Lim\inst{1} \and Alexander Dielen\inst{1} \and Marcel Campen\inst{2} \and Leif Kobbelt\inst{1}}

\institute{Visual Computing Institute, RWTH Aachen University \and Osnabr\"uck University}

\maketitle

\begin{abstract}
The question of representation of 3D geometry is of vital importance when it comes to leveraging the recent advances in the field of machine learning for geometry processing tasks. For common unstructured surface meshes state-of-the-art methods rely on patch-based or mapping-based techniques that introduce  resampling operations in order to encode neighborhood information in a structured and regular manner. We investigate whether such resampling can be avoided, and propose a simple and direct encoding approach. It does not only increase processing efficiency due to its simplicity -- its direct nature also avoids any loss in data fidelity. To evaluate the proposed method, we perform a number of experiments in the challenging domain of intrinsic, non-rigid shape correspondence estimation. In comparisons to current methods we observe that our approach is able to achieve highly competitive results.

\end{abstract}

\section{Introduction}
The representation of 3D geometry is a key issue in the context of machine learning in general and deep learning in particular. A variety of approaches, from point clouds over voxel sets to range images, have been investigated. 
When the input geometry is in the common form of a surface mesh, conversion to such representations typically comes with losses in fidelity, accuracy, or conciseness. 
Hence, techniques have been introduced to more or less directly take such discrete surface data as input to machine learning methods. Examples are graph-based \cite{kostrikov2017surface,defferrard2016convolutional} and patch-based approaches \cite{masci2015geodesic,boscaini2016learning,monti2017geometric}.
While graph-based techniques rely on fixed mesh connectivity structures, patch-based techniques provide more flexibility. However, they crucially rely on some form of (re)sampling of the input mesh data, so as to achieve consistent, regular neighborhood encodings, similar to the regular pixel structures exploited for learning on image data.

In this paper we consider the question whether such resampling can be avoided, taking the mesh data as input even more directly. The rationale for our interest is twofold: the avoidance of resampling would increase the efficiency of inference (and perhaps training) and could possibly increase precision. The increase in efficiency would be due to not having to perform the (typically non-trivial) resampling (either as a preprocess or online). One could hypothesize an increase in precision based on the fact that resampling is, in general, accompanied by some loss of data fidelity.

We propose a resampling and conversion free input encoding strategy for local neighborhoods in manifold 3D surface meshes. In contrast to many previous approaches for learning on surface meshes, we then make use of RNNs and fully-connected networks instead of CNNs, so as to be able to deal with the non-uniform, non-regular structure of the input.
Though simple, this raw input encoding is rich enough that our networks could, in theory, learn to emulate common patch resampling operators based on it. Nevertheless, hand-crafting such resampling operators and preprocessing the input accordingly, as previously done, could of course be of benefit in practice. Hence it is important to evaluate practical performance experimentally.

We apply and benchmark our technique in the context of \emph{non-rigid shape correspondence estimation} \cite{van2011survey}. 
The computation of such point-to-point (or shape) correspondences is of interest for a variety of downstream shape analysis and processing tasks (e.g. shape interpolation, texture transfer, etc.). 
The inference of these correspondences, however, is a challenging task and topic of ongoing investigation.
Our experiments in this context reveal that the preprocessing efforts can indeed be cut down significantly by our approach without sacrificing precision. In certain scenarios, as hypothesized, precision can even be increased relative to previous resampling-based techniques.

\paragraph{\textbf{Contribution}} In this work we propose and investigate a novel form of using either fully-connected layers or LSTMs (Hochreiter and Schmidhuber~\cite{hochreiter1997long}) for point-to-point correspondence learning on manifold 3D meshes. 
By serializing the local neighborhood of vertices we are able to encode relevant information in a straightforward manner and with very little preprocessing. 
We experimentally analyze the practical behavior and find that our approach achieves competitive results and outperforms a number of current methods in the task of shape correspondence prediction.

\section{Related Work}
\begin{figure}[tb]
\centering
\subfloat[][]{
\begin{overpic}[width=0.4\textwidth]
{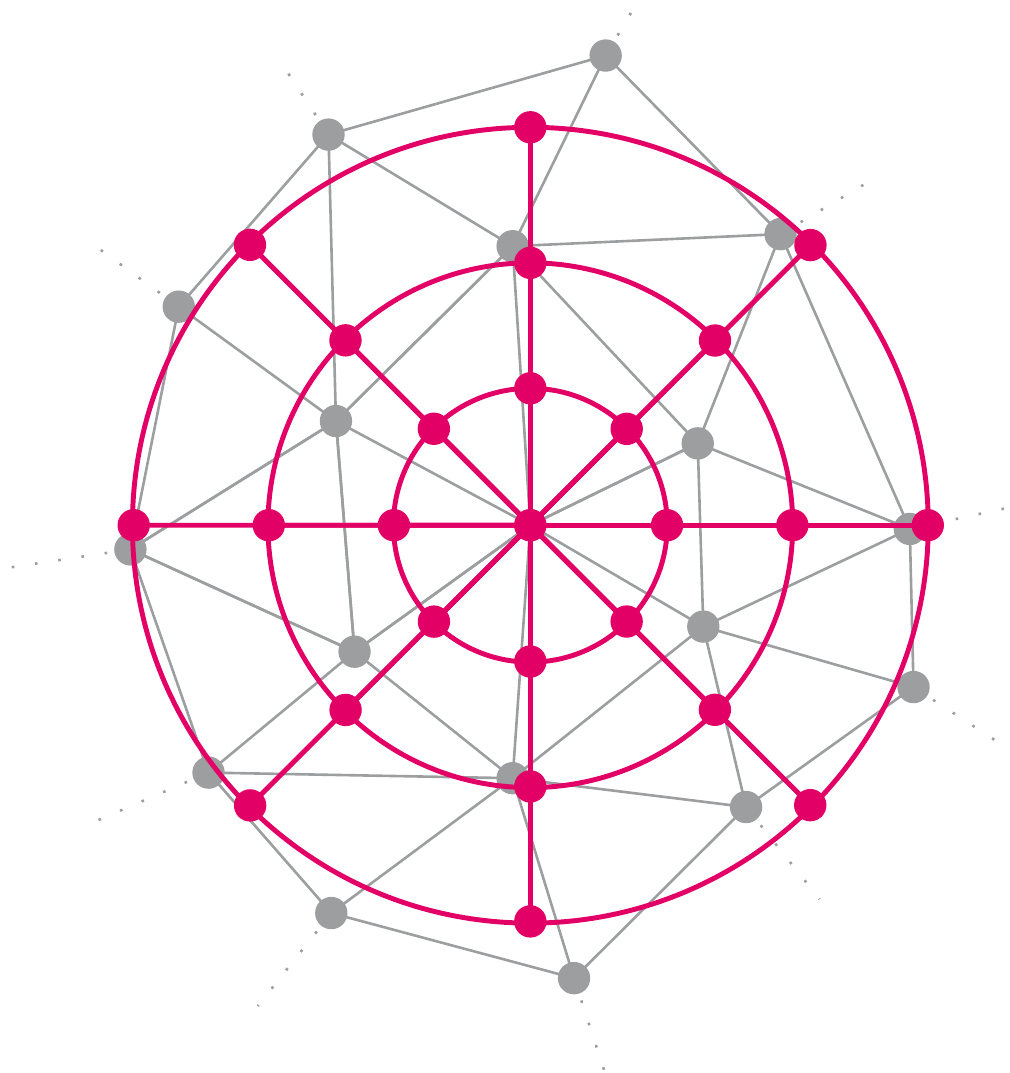}
\put(80,75){\color[RGB]{227,0,102}$(r,\theta)$}
\end{overpic}
\label{fig:patch_geod}
}
\subfloat[][]{
\begin{overpic}[width=0.4\textwidth]
{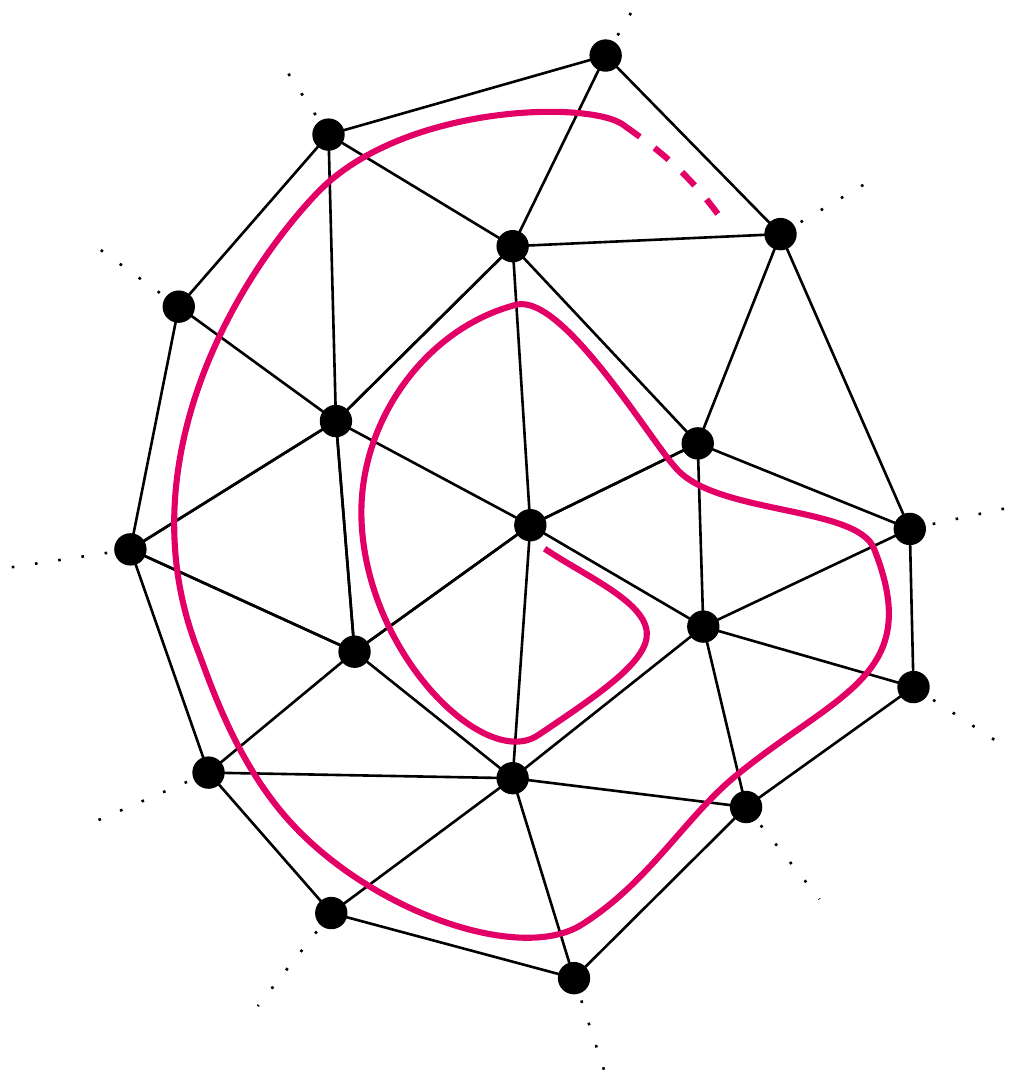}
\put(52,50){$a$}
\put(70,35){$b$}
\put(70,60){$g$}
\put(52,80){$f$}
\put(25,60){$e$}
\put(32,32){$d$}
\put(52,22){$c$}
\end{overpic}
\label{fig:patch_mesh}
}
\caption{The black graph represents a patch of a triangle mesh. 
(a) For generalized CNNs on 3D meshes~\cite{masci2015geodesic,boscaini2016learning,monti2017geometric}, we would have to compute a blended $\mathrm{f}(r,\theta)$ for each node of the magenta polar grid in order to provide a fixed number and pattern of samples for a convolution kernel.
(b) Instead, we enumerate the neighborhood vertices of a center vertex $a$ by following a spiral pattern (magenta). For a given feature $\mathrm{f}(\cdot)$ we encode the local neighborhood information feeding $[\mathrm{f}(a),\allowbreak\mathrm{f}(b),\allowbreak\mathrm{f}(c),\allowbreak\mathrm{f}(d),\allowbreak\mathrm{f}(e),\allowbreak\mathrm{f}(f),\allowbreak\mathrm{f}(g), \ldots]$ into a LSTM Cell.
}
\label{fig:patch}
\end{figure}
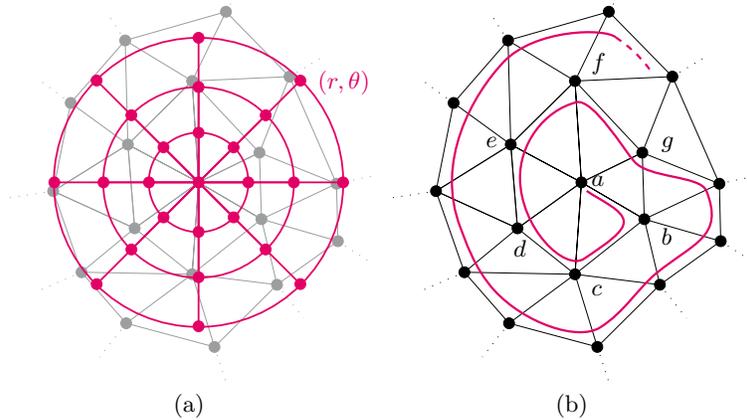

Several data- and model-driven approaches for finding correspondences between shapes have been proposed in previous works.
\paragraph{\textbf{Functional Maps}}
Ovsjanikov et al.~\cite{ovsjanikov2012functional} approach the problem of finding point-to-point correspondences by formulating a function correspondence problem.
They introduce functional maps as a compact representation that can be used for point-to-point maps. Various (model- and data-driven) improvements have been suggested \cite{kovnatsky2013coupled,pokrass2013sparse,huang2014functional,eynard2015multimodal,eynard2016coupled,rodola2017partial,nogneng2017informative,nogneng2018improved,Gehre:2018:InteractiveFunctionalMaps}.
Most closely related to our approach, Litany et al.~\cite{litany2017deep} use deep metric learning to optimize input descriptors for the functional maps framework.
However, point-to-point correspondence inference in all cases requires the computation of a functional map for each pair of shapes. This possibly costly computation can be avoided with our approach. Once trained, our model can be applied directly for inference.
\paragraph{\textbf{Generalized CNNs for 3D Meshes}}
Several data-driven methods that do not rely on functional maps were proposed in recent years.
Masci et al.~\cite{masci2015geodesic} generalize convolution operations in modern deep learning architectures to non-Euclidean domains.
To this end they define geodesic disks (patches) around each vertex. Based on a local polar coordinate system the patches can be resampled with a fixed number and fixed pattern of samples (cf.\ Figure \ref{fig:patch_geod}). This predefined sampling pattern allows to construct a convolution operation on these patches by computing weighted sums of features at sample positions.
In order to transfer the information (i.e. descriptors) available discretely at the vertices to the continuous setting of the geodesic disks for the purpose of resampling, they are blended by means of appropriate kernels.
Boscaini et al.~\cite{boscaini2016learning} propose to use anisotropic kernels in this context, while aligning the local coordinate systems with the principal curvature directions.
Monti et al.~\cite{monti2017geometric} generalize the construction of these blending kernels to Gaussian Mixture Models, which avoids the  hand-crafting of kernels in favor of learning them.

Ezuz et al.~\cite{ezuz2017gwcnn} and Maron et al.~\cite{maron2017convolutional} both propose forms of global (instead of local patch-wise) structured resampling of the surface,
which can then be used as input to well-known CNN architectures used in computer vision.

Similar in spirit to our work is the method introduced by Kostrikov et al.~\cite{kostrikov2017surface}. They apply Graph Neural Networks (cf.~\cite{scarselli2009graph,defferrard2016convolutional,niepert2016learning}) in the domain of 3D meshes.
A key difference is that their network's layers see neighborhood information in reduced blended form (via Laplace or Dirac operators) rather than natively like our approach.

In comparison to these approaches we require very little preprocessing, no heavy online computation, and no resampling. Per-vertex descriptors are exploited directly rather than taking blended versions of them as input.

\section{Resampling-free Neighborhood Encoding}

We assume that the input domain is represented as a manifold triangle mesh $\mathcal{M}$. 
Some form of input data (e.g. positions, normals, or geometry descriptors) is specified or can be computed at the vertices of $\mathcal{M}$. We denote the information (\emph{feature}) at a vertex $v$ by $\allowbreak\mathrm{f}(v)$.
As in previous work \cite{masci2015geodesic,boscaini2016learning,monti2017geometric}, for the task of correspondence estimation, we would like to collect this information $\mathrm{f}$ from a local neighborhood around a vertex $a$. As mentioned above, we intend to encode this relevant information in a very direct manner, essentially by a notion of serialization of the per-vertex features $\mathrm{f}$ in local neighborhoods, without any alterations.
 
 \subsection{Spiral Operator}
To this end we make the observation that, given a center vertex, the surrounding vertices can quite naturally be enumerated by intuitively following a spiral, as illustrated in Figure \ref{fig:patch_mesh}. The only degrees of freedom are the orientation (clockwise or counter-clockwise) and the choice of 1-ring vertex marking the spiral's starting direction. We fix the orientation to clockwise here. The choice of starting direction is arbitrary, and a different sequence of vertices will be produced by the spiral operator depending on this choice. This rotational ambiguity is a common issue in this context, and has been dealt with, for instance, by max-pooling over multiple choices \cite{masci2015geodesic}, or by making the choice based on additional, e.g.\ extrinsic, information \cite{boscaini2016learning}.
We avoid this by instead making a random choice in each iteration during training, enabling the network to learn to be robust against this ambiguity, assuming a sufficient number of parameters in the network.

Given a starting direction (i.e. a chosen 1-ring vertex), the spiral operator produces a sequence enumerating the center vertex, followed by the 1-ring vertices, followed by the 2-ring vertices, and so forth. Thus, 
for a given $k$, it is possible to trace the spiral until we have enumerated all vertices up to and including the $k$-ring.
In Figure~\ref{fig:patch_mesh} this is illustrated for the case $k=2$, where the sequence reads $[a,b,c,d,e,f,g,\ldots]$. Alternatively, for a given $N$, we can of course trace until we have enumerated exactly $N$ vertices, thereby producing fixed length sequences -- in contrast to the variable length sequences up to ring~$k$.

While the definition and practical enumeration of a spiral's vertices is really simple locally, some care must be taken to support the general setting, in particular with large $k$ or large $N$ (when $k$-rings are not necessarily simple loops anymore) or on meshes with boundary (where $k$-rings can be partial, maybe consisting of multiple components). The following concise definition of the spiral operator handles also such cases.

Let $k$-ring and $k$-disk be defined as follows:
\begin{align*}
    0\text{-ring}(v) &= \{v\}, \\
    (k\!+\!1)\text{-ring}(v) &= N(k\text{-ring}(v)) \,\backslash\, k\text{-disk}(v),\\
    k\text{-disk}(v) &= \cup_{i = 0 \dots k}\, i\text{-ring}(v),
\end{align*}
where $N(V)$ is the set of all vertices adjacent to any vertex in set $V$.

The spiral$(v, k)$ is defined simply as the concatenation of the \emph{ordered} rings:
\begin{align*}
    \text{spiral}(v, k) &= (0\text{-ring}(v)  \,\dots\,  k\text{-ring}(v)).
\end{align*}
The fixed-length spiral$(v,N)$ is obtained by truncation to a total of $N$ vertices.

The required order $<$ on the vertices of a $k$-ring is defined as follows:
The 1-ring vertices are ordered clockwise, starting at a random position. The
ordering of the $(k\!+\!1)$-ring vertices is induced by their k-ring neighbors in the
sense that vertices $v_1$ and $v_2$ in the $(k\!+\!1)$-ring being adjacent to a
common vertex $v^{*}$ in the $k$-ring are ordered clockwise around $v^{*}$,
while vertices $v_1$ and $v_2$ having no common k-ring neighbor are sorted
in the same order as (any of) their $k$-ring neighbors.

\subsection{Learning}

With the (either variable length or fixed length) vertex sequence $[a, \allowbreak b,\allowbreak c,\allowbreak d,\allowbreak e,\allowbreak f,\allowbreak g, \dots]$ produced for a given center vertex, one easily serializes the neighborhood features as the sequence $[\mathrm{f}(a),\allowbreak\mathrm{f}(b),\allowbreak\mathrm{f}(c),\allowbreak\mathrm{f}(d),\allowbreak\mathrm{f}(e),\allowbreak\mathrm{f}(f),\allowbreak\mathrm{f}(g), ...]$.

For the purpose of correspondence estimation our goal is to learn a compact high-level representation of these sequences. This can be done in a straightforward and intuitive way using recurrent neural networks. More specifically, we feed our vertex sequences into an LSTM cell as proposed by Hochreiter and Schmidhuber \cite{hochreiter1997long} and use the last cell output as representation. This representation is thus computed using the following equations:
\begin{align*}
f_t &= \sigma(W_f \cdot [x_t,h_{t-1}] + b_f), \\
i_t &= \sigma(W_i \cdot [x_t,h_{t-1}] + b_i), \\
o_t &= \sigma(W_o \cdot [x_t,h_{t-1}] + b_o), \\
c_t &= f_t \odot c_{t-1} + i_t \odot \tanh(W_c \cdot [x_t,h_{t-1}] + b_c), \\
h_t &= o_t \odot \tanh(c_t),
\end{align*}
where the learnable parameters are the matrices $W_f,W_i,W_o,W_c$ with their respective biases $b_f,b_i,b_o,b_c$.
$[x_t,h_{t-1}]$ is the concatenation of the input $x_t$ (e.g. $\mathrm{f}(a)$) and the previous hidden state $h_{t-1}$, while $c_t$ and $h_t$ are the current cell- and hidden-state respectively.
We denote the Hadamard product as $\odot$.

This generation of a representation of the local neighborhood of a vertex via a LSTM cell is, in an abstract sense, comparable to the generalized convolution operation of previous patch-based approaches. However, the resampling of neighborhoods and computation of blended features $\mathrm{f}(r,\theta)$ for each sample $(r,\theta)$ (see Figure~\ref{fig:patch_geod}) is avoided by our approach.
Here $r$ and $\theta$ are geodesic polar coordinates of some local coordinate system located at each center vertex.
$\mathrm{f}(r,\theta)$ is then computed based on a weighted combination of $\mathrm{f}$ at nearby vertices (e.g. $\mathrm{f}(r,\theta)=w_c \mathrm{f}(c) + w_d \mathrm{f}(d) + \cdots)$. Depending on the nature of $\mathrm{f}$ this linear blending can be lossy.

For the case of a fixed length serialization, the use of an RNN supporting variable length input is not necessary. A fully-connected layer (combined with some non-linearity) can be used instead.
Naturally, we apply these neighborhood encoding operations repeatedly in multiple layers in a neural network to facilitate the mapping of input features to a higher level feature representation. This is detailed in the following section.

\paragraph{\textbf{Tessellation Dependence}}

Our simple method of encoding the neighborhood obviously is not independent of the tessellation of the input.
By augmenting the features $\mathrm{f}$ with metric information (i.e. by appending length and angle information), we can mitigate this and essentially enable the network to possibly \emph{learn} to be independent. In Section \ref{sec:rem} we investigate the effects of this.

Concretely, we concatenate to the input feature $\mathrm{f}(c)$ the distance of the current vertex $c$ to the center vertex $a$
as well as the angle at $a$ between the previous vertex $b$ and $c$.

\subsection{Architecture Details}
To evaluate and compare our proposed methods (with variable or fixed length sequences) in the context of shape correspondence estimation, we construct our network architectures in a manner similar to the GCNN3 model proposed by Masci et al.~\cite{masci2015geodesic}. We replace the convolution layers in GCNN3 by the ones presented above, as detailed below.
For the sake of comparability, we use the SHOT descriptor proposed by Salti et al.~\cite{salti2014shot} with 544 dimensions and default parameter settings computed at each vertex as input, following \cite{boscaini2016learning,monti2017geometric}.

The original GCNN3 \cite{masci2015geodesic} network is constructed as FC16 + GC32 + GC64 + GC128 + FC256 + FC6890. FC$x$ refers to a fully connected layer with output size $x$, which is applied to each vertex separately. GC$x$ is the geodesic convolution operation followed by angular max-pooling, producing $x$-dimensional feature vectors for every vertex.

\paragraph{\textbf{LSTM-NET}} Our network (LSTM-NET) for sequences with varying length replaces the GC layers and is constructed as FC16 + LSTM150 + LSTM200 + LSTM250 + FC256 + FC6890. LSTM$x$ is the application of a LSTM cell to a sequence consisting of the input vertex and its neighborhood. In this manner we compute a new feature vector with dimensionality $x$ (encoding neighborhood information) for every vertex, similar to a convolution operation.

\paragraph{\textbf{FCS-NET}} For fixed-length sequences we make use of a network (FCS-NET) constructed as FC16 + FCS100 + FCS150 + FCS200 + FC256 + FC6890.
FCS$x$ refers to a fully-connected layer, which takes the concatenated features of a sequence as input and produces a $x$-dimensional output for every vertex, analogously to the LSTM$x$ operation above.

We apply ReLU~\cite{nair2010rectified} to all layer outputs except for the output of the final layer to which we apply softmax. As regularization we apply dropout~\cite{srivastava2014dropout} with $p=0.3$ after FC16 and FC256. For fair comparison, the layers of our LSTM-NET and FCS-NET were chosen such that the total number of learnable parameters is roughly equal to that of GCNN3 (cf.\ Table~\ref{tab:params}). Our networks are implemented with TensorFlow~\cite{tensorflow2015-whitepaper}.
\begin{table}[tb]
\caption{Number of parameters used in the different network architectures. FCS-NET (20) refers to FCS-NET applied to sequences with length 20, while GCNN3 is our implementation of GCNN3 \cite{masci2015geodesic} with the SHOT descriptor\label{tab:params}}
\centering
\begin{tabular}{ | c | c | }
  \hline
  Network & Number of Parameters\\
  \hline
  GCNN3 (SHOT) & 2,672,634 \\
  LSTM-NET & 2,675,706 \\
  FCS-NET (20) & 2,763,356 \\
  \hline
\end{tabular}
\end{table}

\section{Experiments}

\begin{figure}[b!]
\centering
\subfloat[]{
\includegraphics[width=0.5\textwidth]{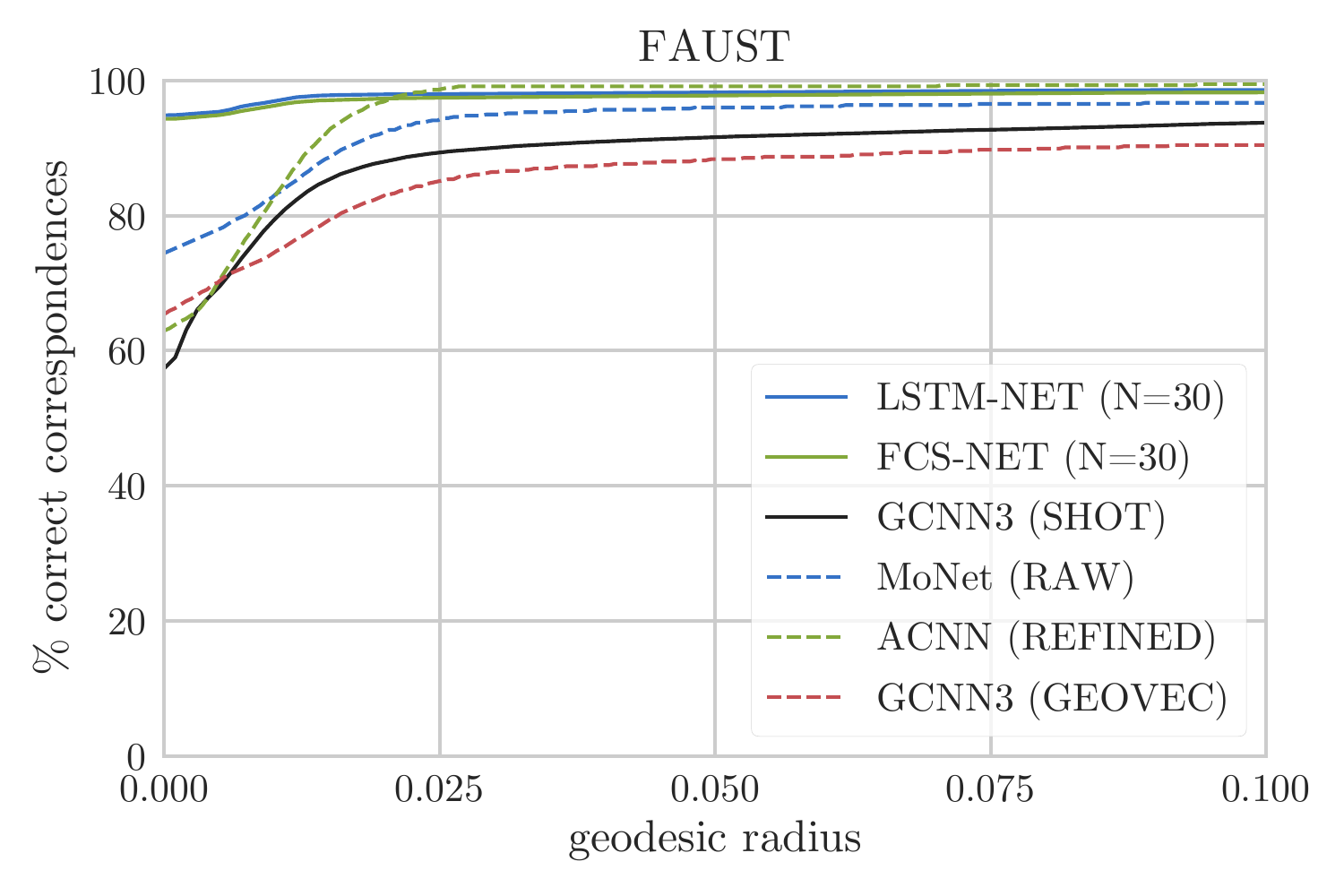}
}
\subfloat[]{
\includegraphics[width=0.5\textwidth]{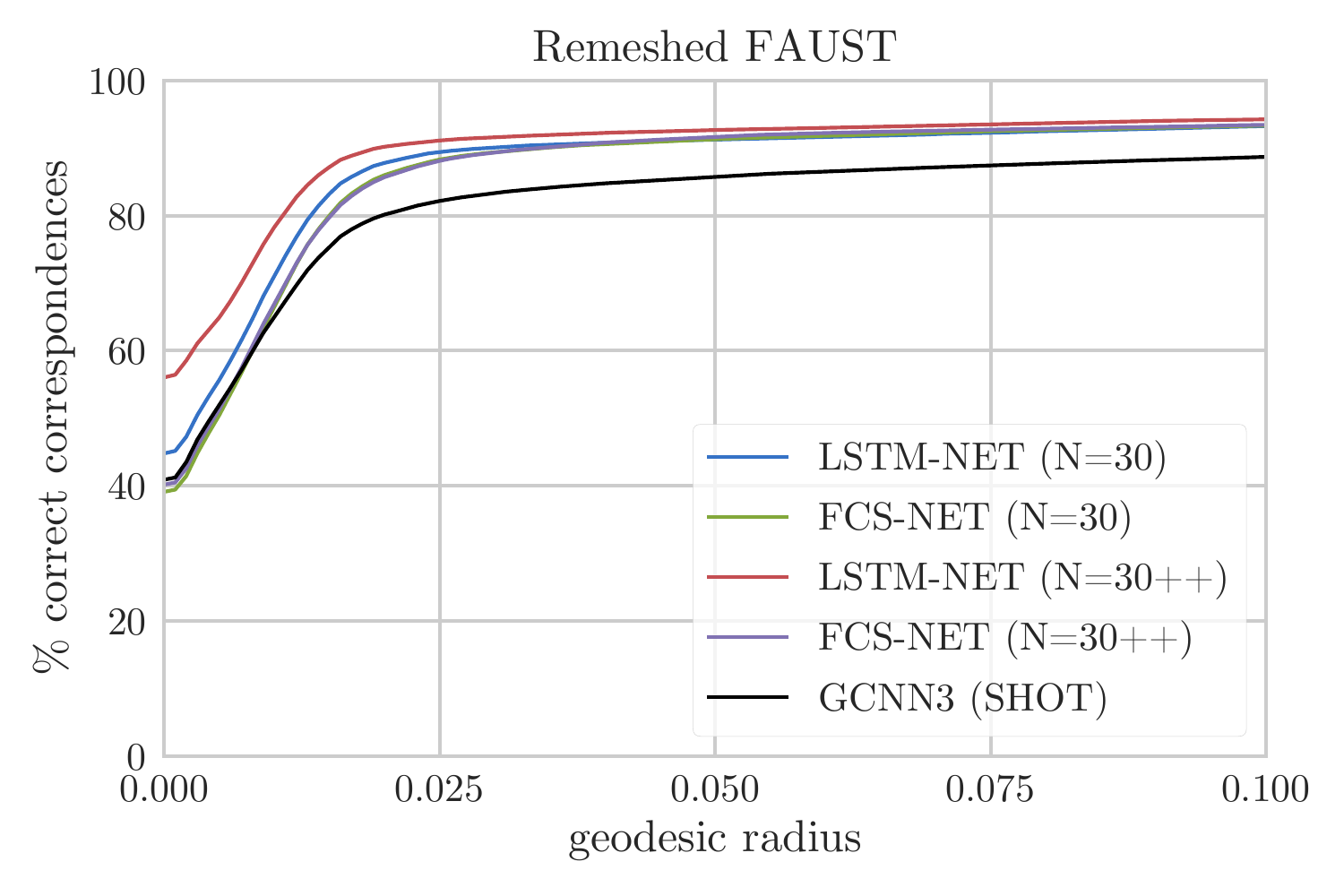}
}
\caption{Here the percentage of correct point-to-point correspondence predictions included in varying geodesic radii is shown. 
(a) shows a comparison of our approaches (FCS-NET, LSTM-NET) on sequences of length N=30 to current approaches. Dashed lines refer to results reported in previous work. For GCNN3 \cite{masci2015geodesic} we compare against the original version that uses the GEOVEC descriptor (dashed) as well as our implementation of GCNN3 (black), which takes the more advanced SHOT descriptor as input. ACNN \cite{boscaini2016learning} shows the results after a correspondence map refinement step. For the sake of fair comparison we show the raw (w/o refinement) performance of MoNet \cite{monti2017geometric}, as we do not perform any refinement for the output of FCS- and LSTM-NET either. (b) visualizes the results on the remeshed FAUST dataset (cf.\ Sec.~\ref{sec:rem}). As expected, the addition of relative angles and distances (++) is beneficial.\label{fig:quantitative}}
\end{figure}
\begin{figure}[tb]
\centering
\begin{minipage}{0.49\textwidth}
\centering
\subfloat[]{
\includegraphics[width=\textwidth]{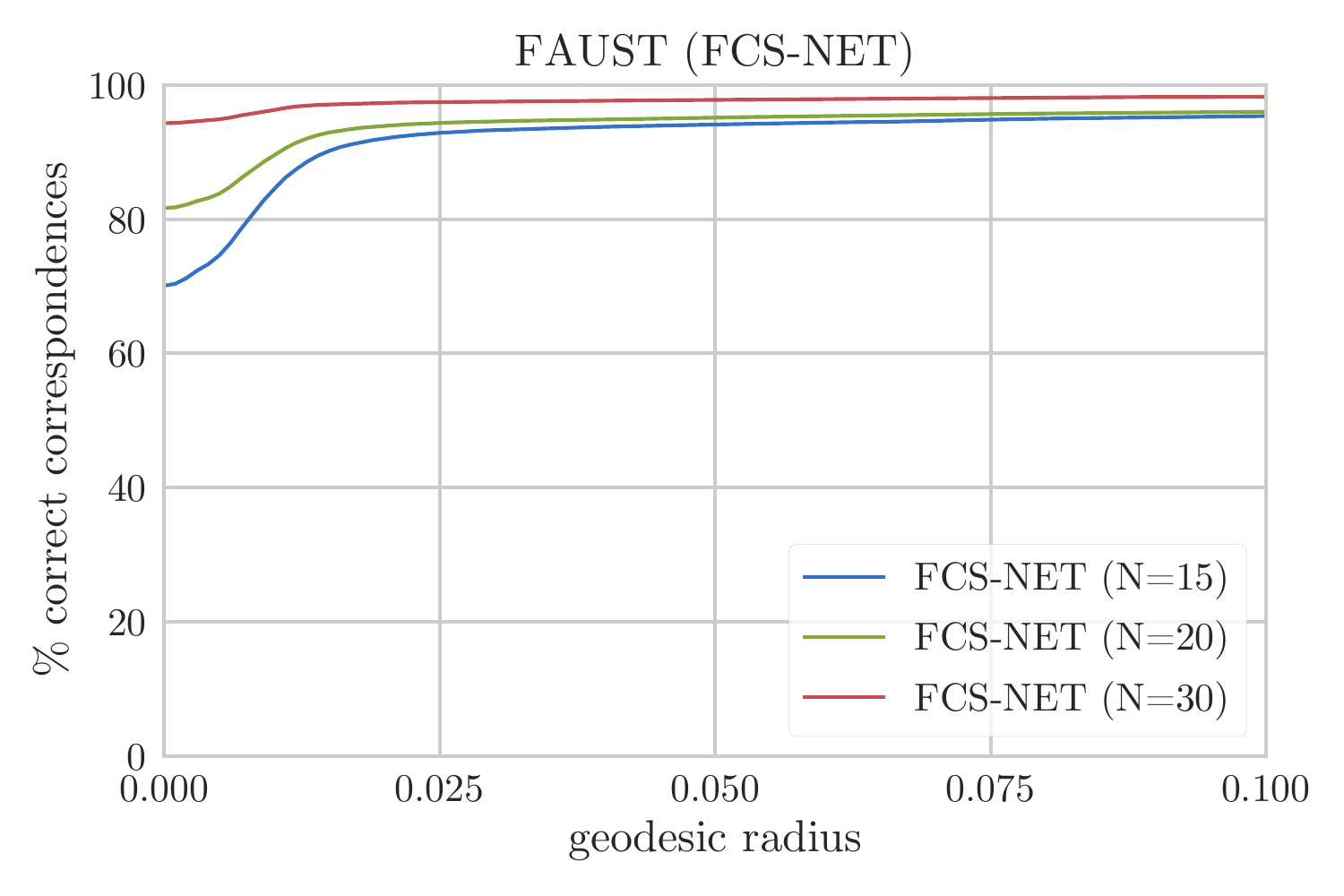}
}\\
\subfloat[]{
\includegraphics[width=\textwidth]{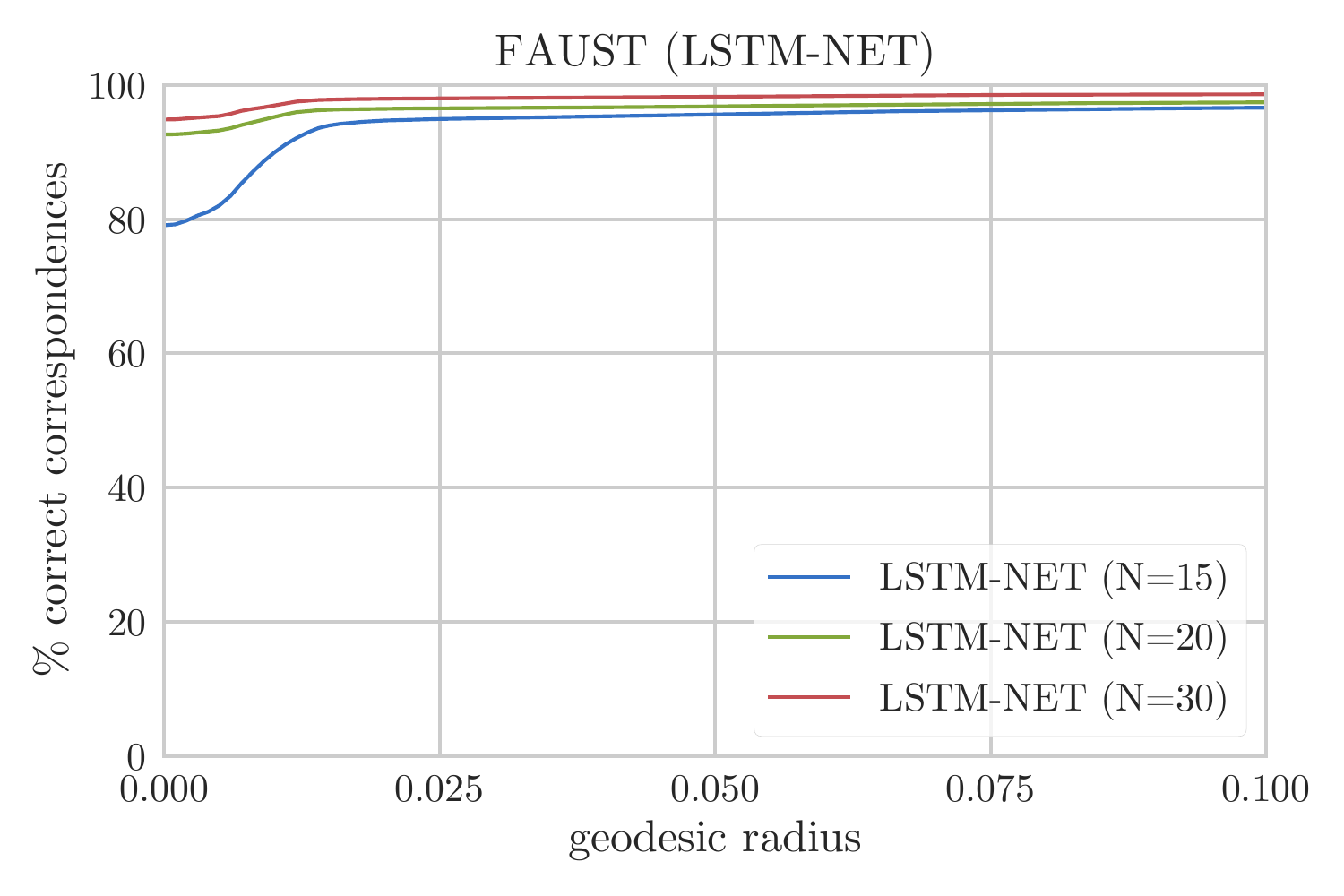}
}
\end{minipage}
\begin{minipage}{0.49\textwidth}
\centering
\subfloat[]{
\includegraphics[width=\textwidth]{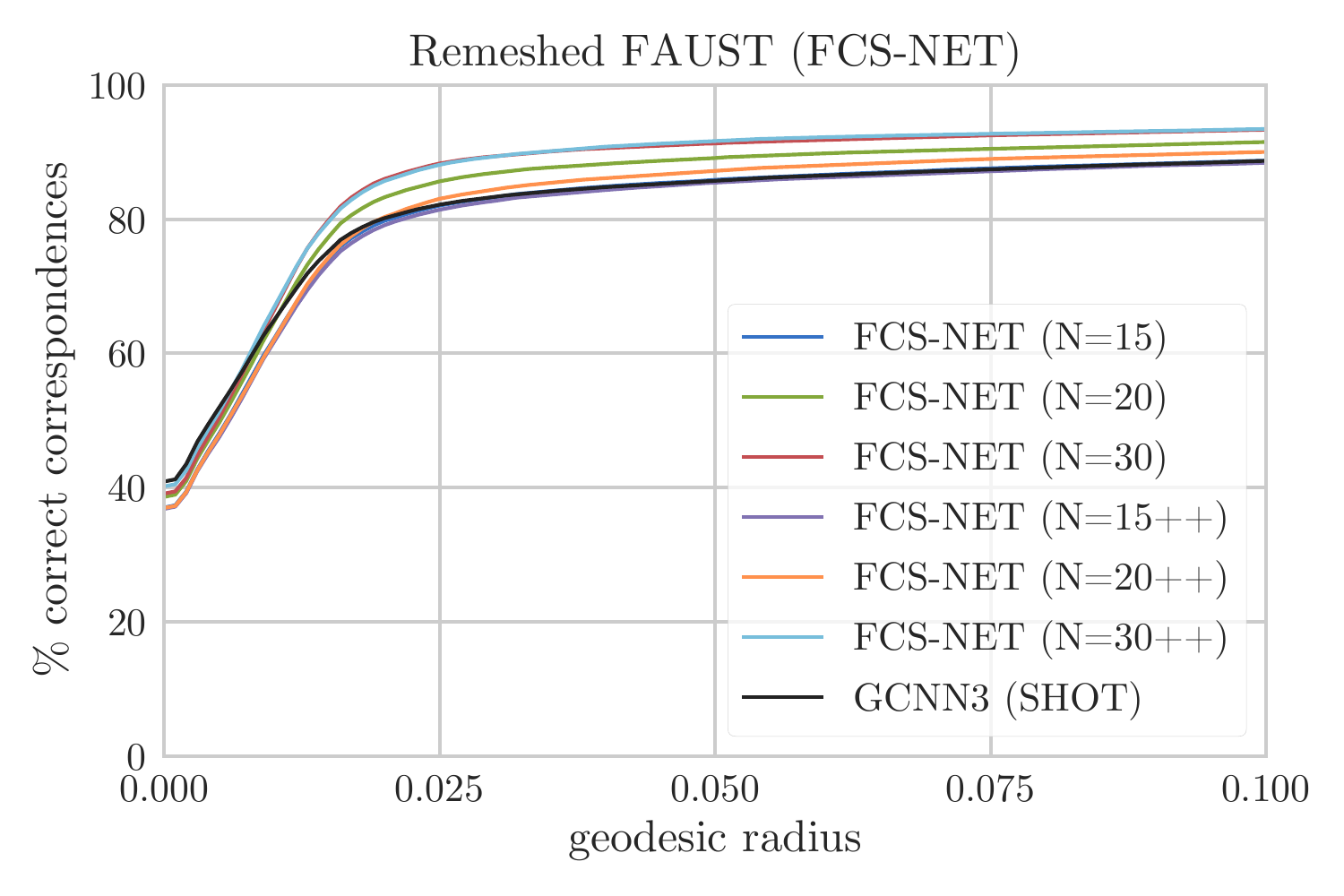}
}\\
\subfloat[]{
\includegraphics[width=\textwidth]{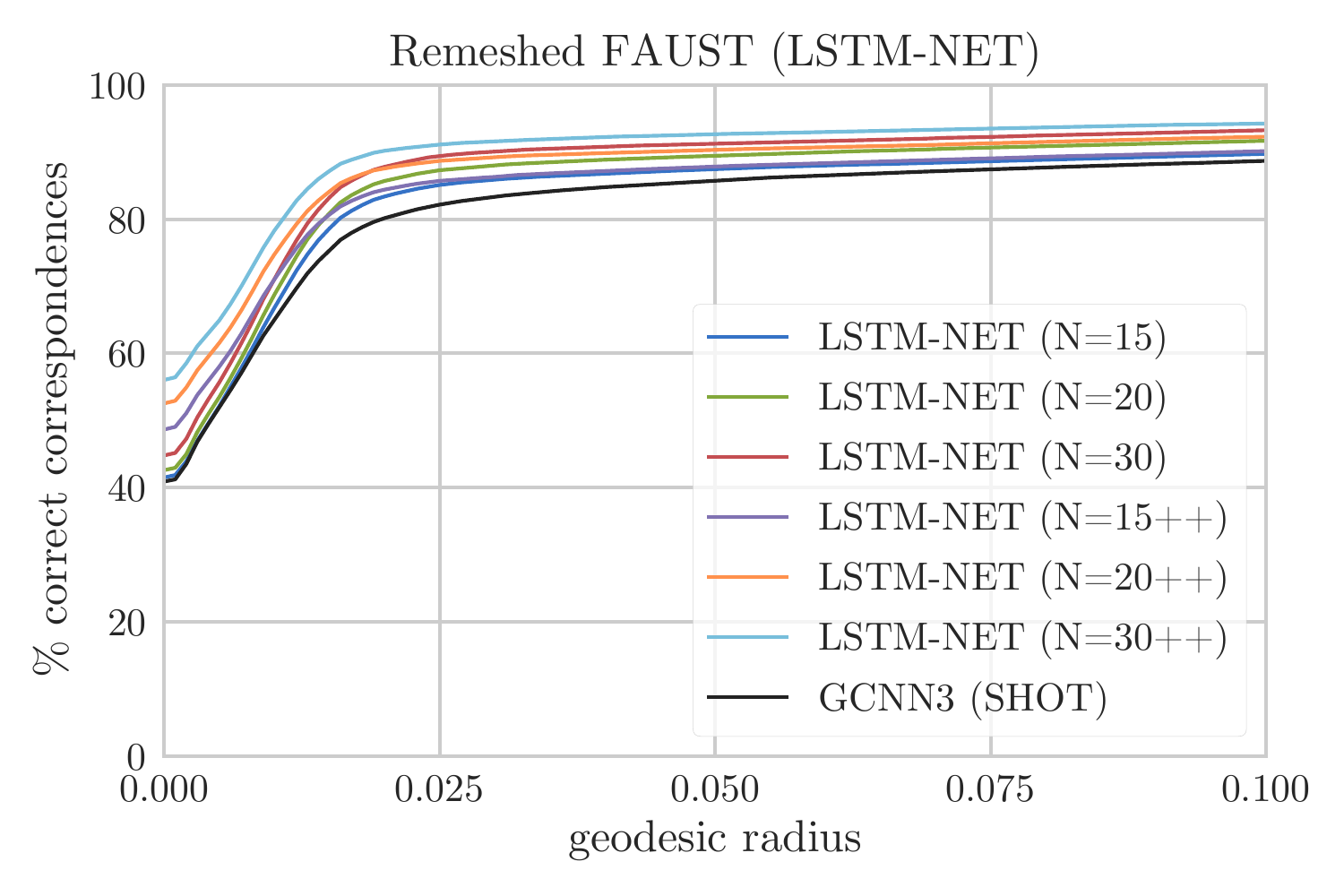}
}
\end{minipage}
\caption{Here the percentage of correct point-to-point correspondence predictions included in varying geodesic radii is shown. 
(a-b) show the effect of different sequence lengths (N=15,20,30) for the FAUST dataset. Even with relatively short sequences (15) we achieve competitive results. (c-d) visualize the results on the remeshed FAUST dataset. For comparison we also show the performance of the GCNN3~\cite{masci2015geodesic} network with the SHOT descriptor. (++) denotes the usage of additional metric  information.\label{fig:quantitative_2}}
\end{figure}

For our experiments we used the FAUST dataset (consisting of 100 shapes)~\cite{Bogo:CVPR:2014}. This allows for comparisons to related previous methods, which have commonly been evaluated on this dataset. Following common procedure, for training we used the first 80 shapes (10 of which were used for validation). All experiment results were computed on the last 20 shapes (our test set). 
We optimized all networks with Adam~\cite{kingma2014adam} ($lr = 0.001$, $\beta_1 = 0.9$, $\beta_2=0.999$), where each batch consisted of the vertices of one mesh.

In order to evaluate the performance of our LSTM-NET we restrict ourself to sequences of fixed length as input (even though it would be capable of dealing with variable length input). This is because the mesh connectivity is the same over all meshes of the dataset. For varying length sequences (e.g. the 1- and 2-ring of each vertex) the network would potentially be able to learn the valence distribution and use connectivity information as an (unfair) prediction help. 

\begin{figure}[tb]
\centering
\begin{overpic}[width=\textwidth]
{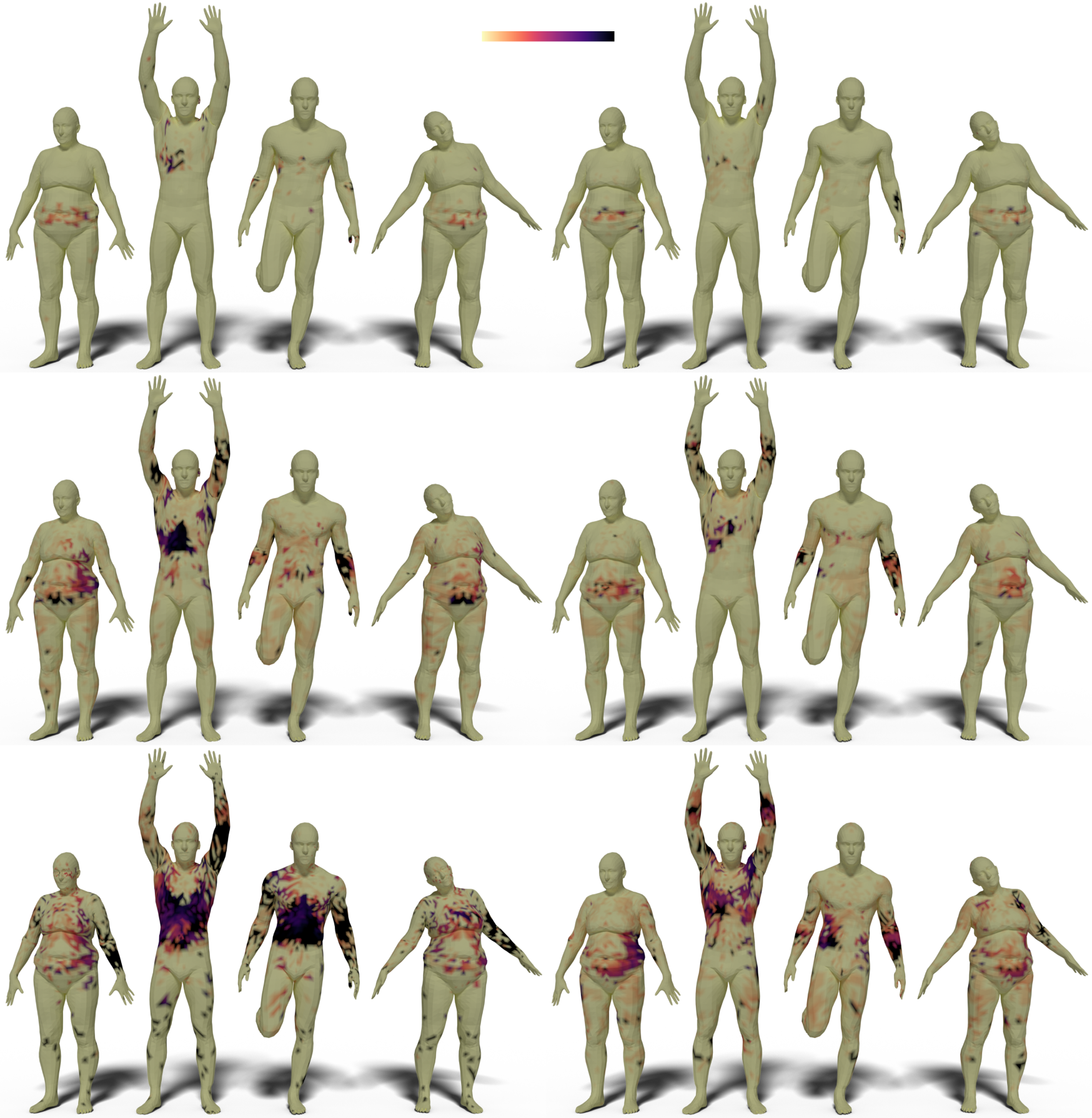}
\put(39.75,97){0.0}
\put(55,97){0.2}
\put(25,94){FCS-NET (30)}
\put(74,94){LSTM-NET (30)}
\put(25,61){FCS-NET (15)}
\put(74,61){LSTM-NET (15)}
\put(25,27){GCNN3 (GEOVEC)}
\put(74,27){GCNN3 (SHOT)}
\end{overpic}
\caption{Geodesic error for 4 shapes from the test set of the FAUST dataset.\label{fig:orig_qual}}
\end{figure}

Following Kim et al.~\cite{kim2011blended} we compute point-to-point correspondences and plot the percentage of correct correspondences found within given geodesic radii. For the evaluation no symmetry information is taken into account. We compare to the results from~\cite{masci2015geodesic,boscaini2016learning,monti2017geometric}. In addition we also implemented GCNN3 (using the SHOT instead of the GEOVEC descriptor as input) after Masci et al.~\cite{masci2015geodesic} and evaluated the method in our setting. We used the parameters and loss proposed in the original paper.
As shown in Figure~\ref{fig:quantitative} (a) our method outperforms current patch-based approaches with both LSTM-NET and FCS-NET for a sequence length of 30. Note that, by contrast, the average number of interpolated vertices in a patch for GCNN3 is 80. Furthermore, we do not perform any post-processing or refinement on the network predictions. An evaluation of the effect of different sequence lengths is visualized in Figure~\ref{fig:quantitative_2} (a-b). Even with shorter sequence lengths (15) our method achieves competitive results.
Qualitative results are visualized in Figure~\ref{fig:orig_qual}. We show the geodesic distance to the ground truth target vertices on four shapes from the test set. Correspondence errors of relative geodesic distance $>0.2$ are clamped for an informative color coding.

\subsection{Tessellation Dependence}
\label{sec:rem}

\begin{figure}[tb]
\centering
\includegraphics[width=0.6\textwidth]{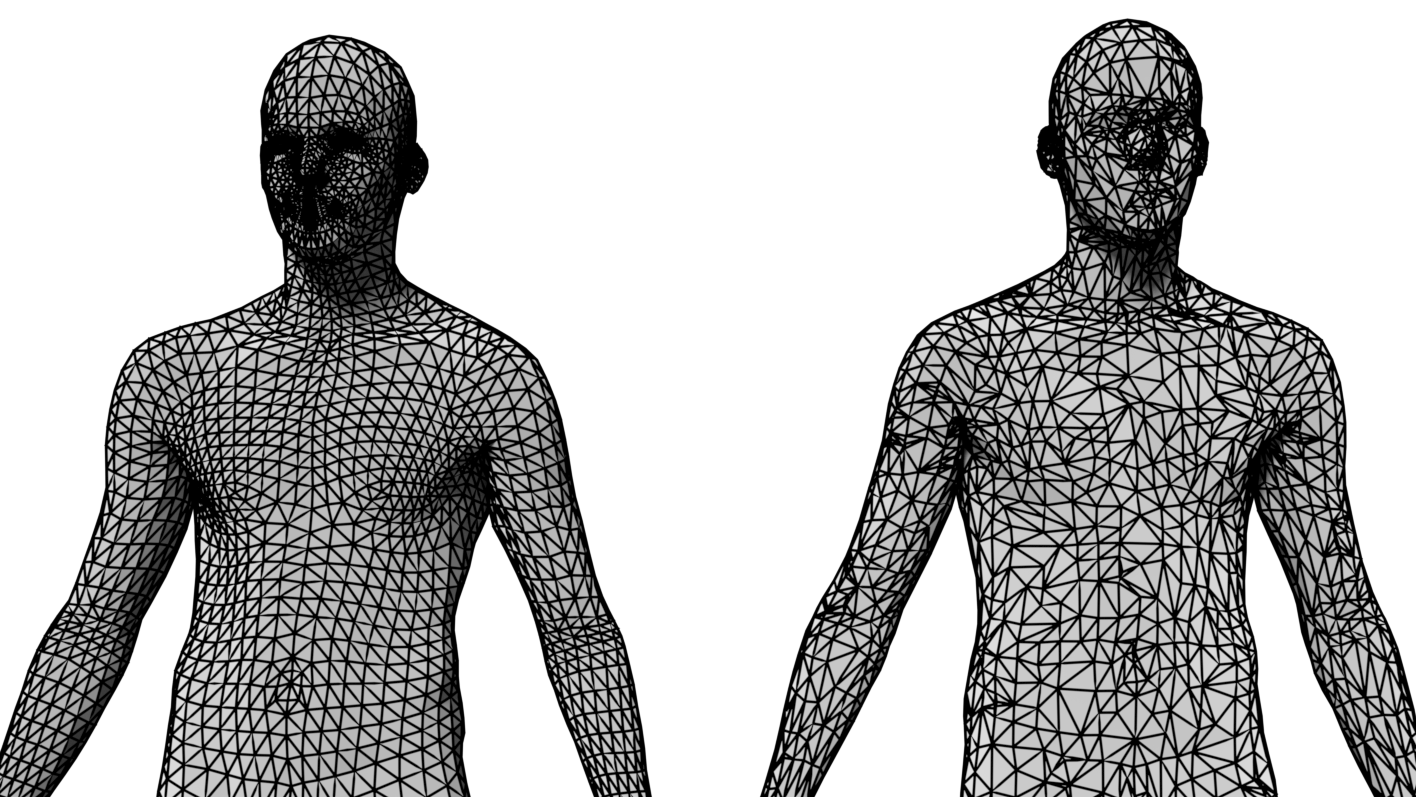}
\caption{Left: triangulation of a shape from the original FAUST dataset. Right: independently remeshed version.}
\label{fig:orig_vs_rem}
\end{figure}

\begin{figure}[tb]
\centering
\subfloat[]{
\includegraphics[width=0.5\textwidth]{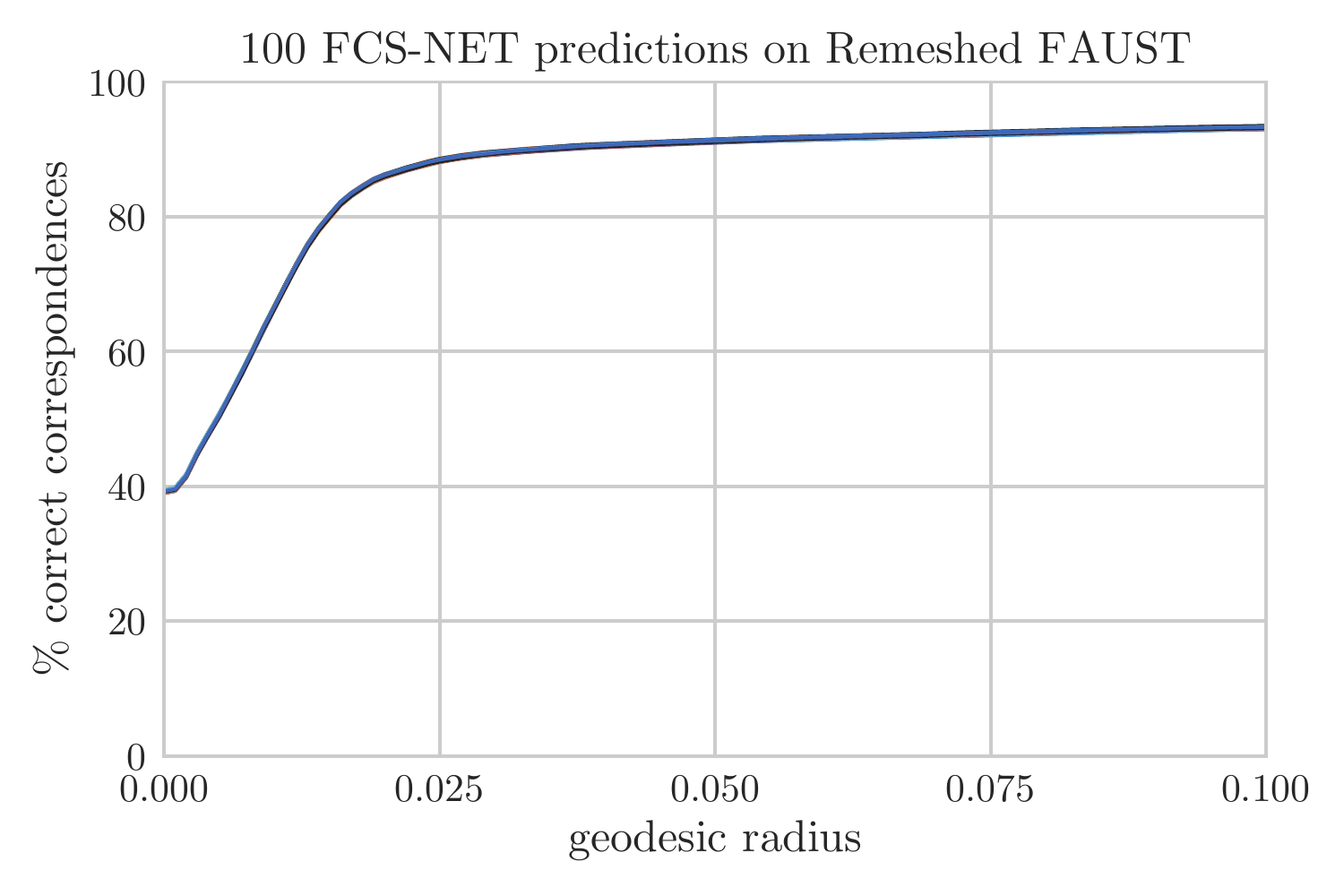}
}
\subfloat[]{
\includegraphics[width=0.5\textwidth]{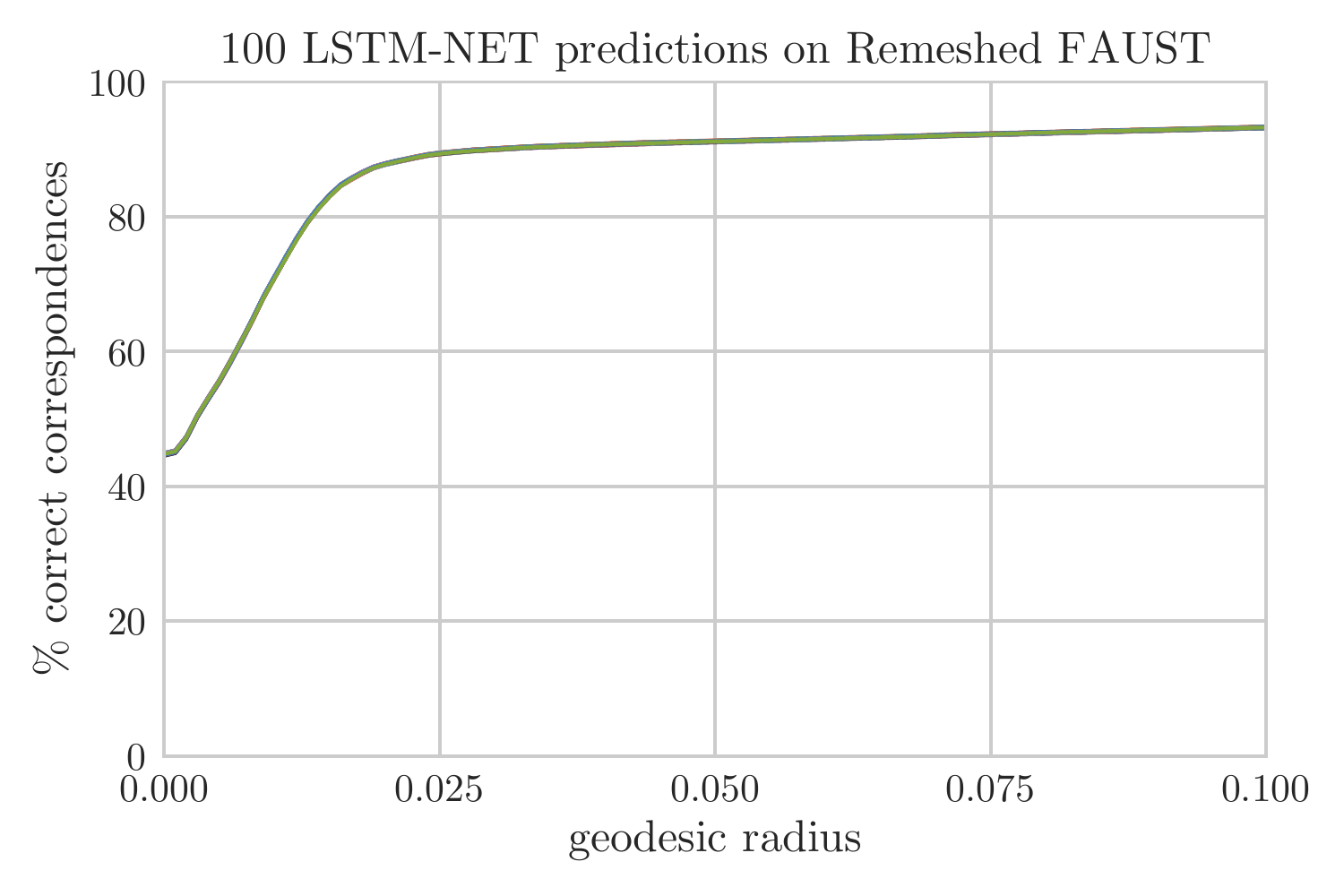}
}
\caption{(a-b) show the robustness of our approach to random rotations of the spirals. We perform 100 inference runs on the test set of the remeshed FAUST dataset with varying random rotations. The 100 different resulting curves plotted here are not distinguishable due to the robustness of our trained networks.\label{fig:100}}
\end{figure}

\begin{figure}[tb]
\centering
\begin{overpic}[width=\textwidth]
{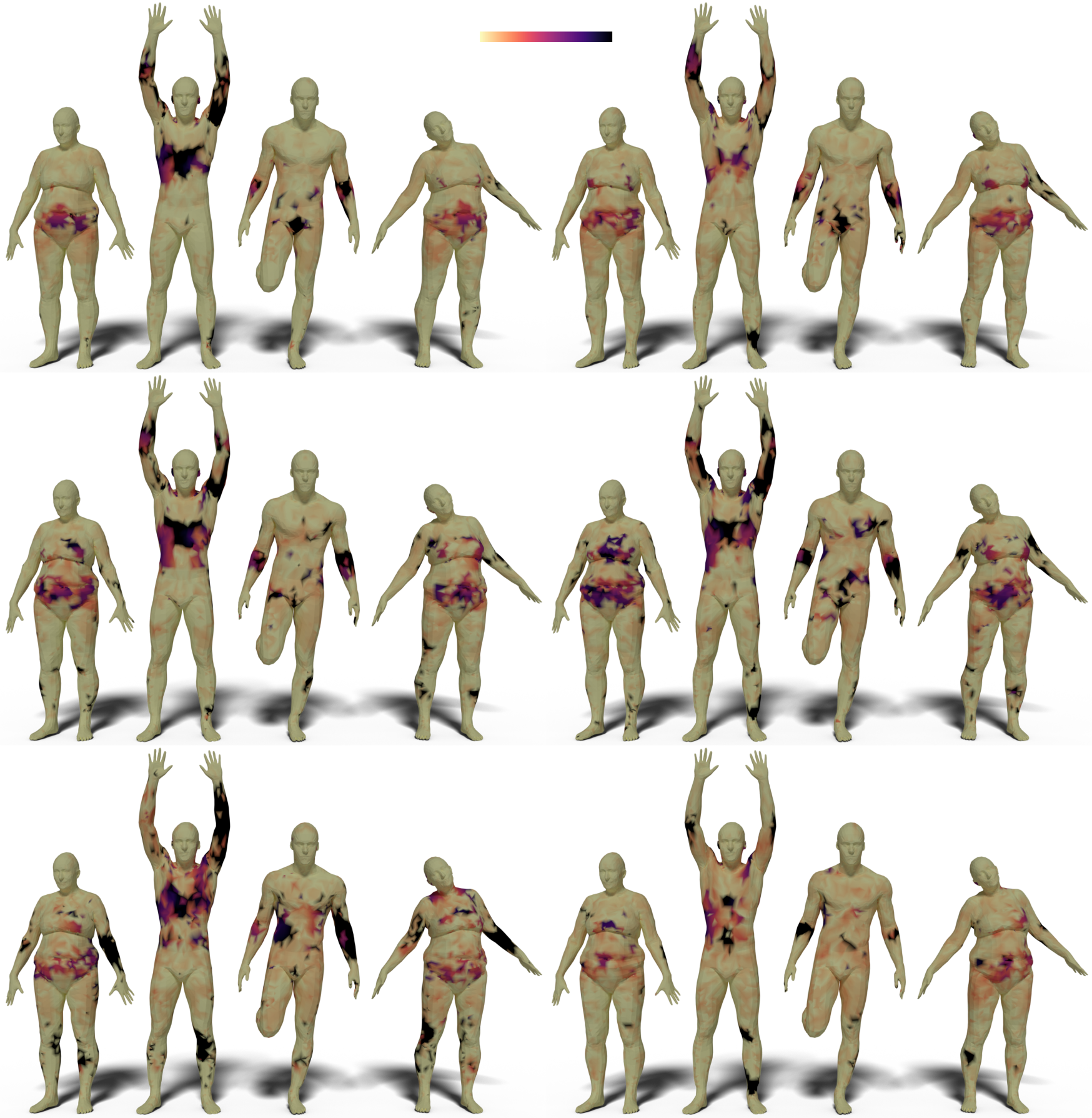}
\put(39.75,97){0.0}
\put(55,97){0.2}
\put(25,94){LSTM-NET (30++)}
\put(74,94){LSTM-NET (30)}
\put(25,61){LSTM-NET (15++)}
\put(74,61){LSTM-NET (15)}
\put(25,27){GCNN3 (SHOT)}
\put(74,27){FCS-NET (30++)}
\end{overpic}
\caption{Geodesic error for 4 shapes from the test set of the remeshed FAUST dataset.
\label{fig:rem_qual}}
\end{figure}

An important, but often overlooked detail is the fact that the shapes in the FAUST dataset are meshed compatibly, i.e. the mesh connectivity is identical across shapes, and identical vertices are at corresponding points. Unless a correspondence estimation method is truly tessellation-oblivious, this naturally has the potential to incur a beneficial bias in this artifical benchmark, as in any realistic correspondence estimation application scenario, the tessellation will of course be incompatible. We thus repeat our experiments with a remeshed version of the FAUST dataset (see Figure~\ref{fig:orig_vs_rem}), where each shape was remeshed individually and incompatibly.

Quantitative results are shown in Figure~\ref{fig:quantitative} (b). Here (++) denotes the additional relative information that we concatenate to the SHOT descriptor vectors. On this more challenging dataset we likewise achieve competitive results.
Especially the additional information (++) enables our networks to encode less tessellation-dependent representations of neighborhoods for better performance.
The effect of different sequence lengths is shown for this dataset in Figure~\ref{fig:quantitative_2} (c-d). For the sake of comparison to the performance of FCS-NET we also restrict LSTM-NET to sequences of fixed length.
See Figure~\ref{fig:rem_qual} for qualitative results.

Furthermore, we test the robustness of our network predictions to random starting points after the center vertex in our sequences (random rotations of the spiral). To this end we perform 100 predictions with different random rotations on the remeshed FAUST dataset with both FCS-NET and LSTM-NET. As shown in Figure~\ref{fig:100} our networks are highly robust to these random orientations, such that the curves of separate predictions are not discernible.

\section{Conclusion}
In this paper we presented a simple resampling free input encoding strategy for local neighborhoods in 3D surface meshes. Previous approaches rely on forms of resampling of input features in neighborhood patches, which incurs additional computational and implementational costs and can have negative effects on input data fidelity.
Our experiments show that our approach, despite its simple and efficient nature, is able to achieve competitive results for the challenging task of shape correspondence estimation.
\paragraph{\textbf{Limitations and Future Work}}
Although the introduction of metric information aims to make our method less sensitive to tessellation, it is nevertheless affected by it; this, however, is true to some extent in any practical setting for previous patch-based approaches as well. The design of truly tessellation-oblivious encoding strategies is a relevant challenge for future work, as it would relieve the training process from having to \emph{learn} tessellation independence, as required for optimal performance.

Furthermore, high resolution meshes require longer sequences to encode relevant neighborhood information. In the case of FCS-NET this also means an increase in the number of parameters required to learn, which can lead to memory issues. An interesting avenue for future work thus is the investigation of sub-sampled (but not resampled) serialization.

A related issue is that the training of RNNs tends to be slower than that of CNNs.
A possible solution to this problem could be the application of 1D convolutions instead of LSTM cells or fully connected layers.
An investigation into feature learning, given only raw input data (e.g.\ lengths, angles, or positions of mesh elements) instead of preprocessed information like the SHOT descriptor will also be of interest.
\subsection*{Acknowledgements}
The research leading to these results has received funding from the European Research Council under the European Union's Seventh Framework Programme (FP7/2007-2013)/ERC grant agreement n$^\circ$ [340884].
We would like to thank the authors of related work \cite{masci2015geodesic,boscaini2016learning} for making their implementations available, as well as the reviewers for their insightful comments.


\end{document}